\def\year{2019}\relax
\documentclass[letterpaper]{article} 
\usepackage{aaai19}  
\usepackage{times}  
\usepackage{helvet}  
\usepackage{courier}  
\usepackage{url}  
\usepackage{graphicx}  
\usepackage{latexsym}
\usepackage{booktabs}
\usepackage{subcaption}
\usepackage{graphicx}
\usepackage{tabularx}
\usepackage{multirow}
\usepackage{mathtools}
\usepackage{algorithm}
\usepackage{algorithmicx}
\usepackage{algpseudocode}
\usepackage{amsmath,amssymb}

\newcommand{\reals}{\mathbb{R}}
\newcommand{\xx}{\mathbf{x}}
\newcommand{\zz}{\mathbf{z}}

\newcommand{\llll}{\mathbf{l}}

\newcommand{\propP}{\mathcal{P}}

\newcommand{\modelM}{\mathbb{M}}

\usepackage{url}

\begin{document}
%
\title{What Is One Grain of Sand in the Desert? \\ Analyzing Individual Neurons in Deep NLP Models}



\author{
  Fahim Dalvi,\thanks{Authors contributed equally}\textsuperscript{1}
  Nadir Durrani,\footnotemark[1]\textsuperscript{1}
  Hassan Sajjad,\footnotemark[1]\textsuperscript{1} \\
  {\bf \Large 
  Yonatan Belinkov,\textsuperscript{2}
  Anthony Bau,\textsuperscript{2}
 James Glass\textsuperscript{2}
  } \\\\
  \textsuperscript{1}Qatar Computing Research Institute, HBKU Research Complex, Doha 5825, Qatar \\ 
  \textsuperscript{2}MIT Computer Science and Artificial Intelligence Laboratory, Cambridge, MA 02139, USA \\
  \texttt{\{faimaduddin,ndurrani,hsajjad\}@qf.org.qa} \\
  \texttt{\{belinkov,abau,glass\}@mit.edu} \\
}

\maketitle
\begin{abstract}

Despite the remarkable evolution 
of deep neural networks in 
natural language processing (NLP), 
their interpretability remains a challenge. 
Previous work largely focused on what these models learn at the representation level. 
We break this analysis down further 
and study individual dimensions (neurons) in the vector representation learned by end-to-end neural models in NLP tasks. We propose two methods: \emph{Linguistic Correlation Analysis}, based on a supervised method to extract the most relevant neurons with respect to an extrinsic task, and \emph{Cross-model Correlation Analysis}, an unsupervised method to extract salient neurons w.r.t. the model itself.  We evaluate the effectiveness of our techniques by ablating the identified neurons and reevaluating the network's performance for two tasks: neural machine translation (NMT) and neural language modeling (NLM). We further present a
comprehensive analysis of neurons with the aim to address the following questions: i) how localized or distributed are different linguistic properties in the models? ii) are certain neurons exclusive to some properties and not others? iii) is the information more or less distributed in NMT vs.\ NLM? and iv) how important are the neurons identified through the linguistic correlation method to the overall task? Our code is publicly available\footnote{\url{https://github.com/fdalvi/NeuroX}} as part of the NeuroX toolkit \cite{neuroX:aaai19:demo}. 

\end{abstract}

\section{Introduction}
While neural networks have achieved state-of-the-art performance in NLP and other spheres of Artificial Intelligence (AI), their opaqueness remains a cause of concern~\cite{DoshiKim2017Interpretability}. 
Interpreting the behavior of 
neural networks is 
considered 
important for increasing trust in AI systems, providing additional information to decision makers, and assisting ethical decision making \cite{lipton2016mythos}.

Recent work attempted to analyze what 
linguistic information is captured in such models when they are trained on a downstream task like neural machine translation (NMT). 
A typical framework is to generate vector representations for some linguistic unit and predict a property of interest such as morphological features. 
This approach has also been applied for analyzing word and sentence embeddings~\cite{qian-qiu-huang:2016:P16-11,adi2016fine}, 
and hidden states in NMT models~\cite{shi-padhi-knight:2016:EMNLP2016,belinkov:2017:acl}.
The analyses reveal that neural vector representations often contain substantial amount of linguistic information. 
Most of this work, however,
targets the whole vector representation, neglecting 
the individual dimensions in the embeddings. In contrast, much work in computer vision investigates properties encoded in individual neurons or filters~\cite{zeiler2014visualizing,zhou2015cnnlocalization}. 


We 
address this gap 
by studying individual dimensions (neurons) in the vector representations learned by end-to-end neural models. 
We aim to increase model transparency by identifying specific 
dimensions that are responsible for 
particular properties. We thus strive for post-hoc decomposibility, in the sense of \cite{lipton2016mythos}. That is, we analyze models after they 
have been trained, in order to uncover the importance of their individual parameters. 
This kind of analysis is important for improving 
understanding of the inner workings of neural networks. It also has potential applications in model distillation (e.g., by removing unimportant neurons), neural architecture search (by guiding the search with important neurons), and mitigating model bias (by identifying neurons responsible for sensitive attributes like gender, race or politeness\footnote{E.g., controlling the system to generate outputs with the right honorifics (``Sie'' vs.\ ``du'') in German.}). 
In this work 
we lay out a methodology for identifying and analyzing individual neurons, and open the call to explore such use cases to the research community. 

To this end, we propose two methods to facilitate neuron analysis. First, we perform an extrinsic correlation analysis through supervised classification on a number of linguistic properties that are deemed important for the task (for example, learning word morphology lies at the heart of modeling various NLP problems). 
Our classifier extracts important individual (or groups of) neurons that capture certain properties. We call this method \emph{Linguistic Correlation Analysis}. 
Second, we propose an alternative methodology to search for neurons that share similar patterns in independently trained networks, based on the assumption that important properties are captured in multiple networks by individual neurons. We call this method \emph{Cross-model Correlation Analysis}. Such an analysis is more intrinsic and helpful for highlighting 
important neurons for the model itself, and in the case when annotated data (supervision) may not be available.
Both machine translation and language modeling are fundamental AI tasks that have seen tremendous improvements with neural networks in recent years. We evaluated our methods for analyzing 
neurons on these two tasks. 

We provide
quantitative evidence that our rankings are 
correct by performing several ablation experiments: from masking out important neurons to removing them completely from the training. We then conduct a comprehensive analysis of the ranked neurons. Our analysis reveals interesting findings such as i) open class categories such as \emph{verb} (part-of-speech tag) and \emph{location} (semantic entity) are much more distributed across the network compared to closed class categories such as \emph{coordinating conjunction} (e.g., ``but/and'') or a \emph{determiner} (e.g., ``the''), 
ii) the model recognizes a hierarchy of linguistic properties and distributes neurons based on it, and iii) important neurons extracted from the \emph{Cross-model Correlation} method overlap with those extracted from the \emph{Linguistic Correlation} method; for example, both methods identified the same neurons capturing position as salient.  
In summary, we make the following contributions:

\begin{itemize}
	\item A general methodology for identifying linguistically-meaningful neurons in deep NLP models. 
	\item An unsupervised method for finding important neurons in neural networks, and a quantitative evaluation of the retrieved neurons.
	\item Application to various test cases, investigating core language properties through part-of-speech (POS), morphological, and semantic tagging.
	\item An analysis of distributed vs.\  focused information in NMT and NLM models. 
\end{itemize}

\section{Related Work}

Much of the previous work has looked into neural models from the perspective of what they learn about various language properties. This includes analyzing word and sentence embeddings~\cite{adi2016fine,qian-qiu-huang:2016:P16-11,conneau2018you}, recurrent neural network (RNN) states~\cite{shi-padhi-knight:2016:EMNLP2016,wang2017gate}, and NMT representations \cite{belinkov:2017:acl,I17-1001,dalvi-EtAl:2017:I17-1}. 
The language properties mainly analyzed are morphological \cite{qian-qiu-huang:2016:P16-11,vylomova2016word}, semantic \cite{qian-qiu-huang:2016:P16-11} and syntactic \cite{shi-padhi-knight:2016:EMNLP2016,linzen2016assessing,conneau2018you}.

Most of this work used an extrinsic supervised task and 
target entire vector representations. 
We study the individual neurons in the vector representation 
and propose a simple supervised method to analyze individual/groups of neurons 
with respect to various properties and linguistic tasks. As an alternative to supervision which is limited to labeled data, we propose an unsupervised method based on correlation 
between several networks to identify salient neurons. 

Some recent work on neural language models and machine translation analyzes specific neurons of length \cite{qian-qiu-huang:2016:EMNLP2016,D16-1248} and sentiment \cite{radford2017learning}. However, not much work has been done along these lines. We present both intrinsic and extrinsic methods to analyze 
models at the neuron level to gain a deeper insight. 

In computer vision, there has been much work on visualizing and analyzing individual units such as filters  in convolutional neural networks~\cite[among others]{zeiler2014visualizing,zhou2015cnnlocalization}. Even though some doubts were cast on the importance of individual units~\cite{s.2018on}, recent work stressed their contribution to predicting specific object classes via ablation studies similar to the ones we conduct~\cite{zhou2018revisiting}. 


\section{Methodology}
Let $\xx = \{\xx_1, \dots, \xx_n\}$ denote a sequence of input features and consider a neural network model $\modelM$ that maps $\xx$ to a sequence of latent representations: $\xx\xmapsto{\modelM}\zz = \{\zz_1, \dots, \zz_n\}$, where $\zz_i \in \reals^D$. For example, in an NMT system, $\modelM$ could be the \emph{encoder}, $\xx$ the input word embeddings, and $\zz$ the hidden states. Our goal is to study individual neurons in the model $\modelM$, which we define as dimensions in the latent representation. We will use $\zz_{ij}$ to denote the $j$-th dimension of the latent representation  of the $i$-th word $\zz_i$. 
We first explain a \emph{Linguistic Correlation Analysis} method to find neurons specific to a task. Then we present a \emph{Cross-model Correlation Analysis} method for ranking based on the correlations between neurons from different networks.

\subsection{Linguistic Correlation Analysis}
\label{sec:approach-supervised}
\begin{figure}[t]
	\centering
	\includegraphics[scale=0.5]{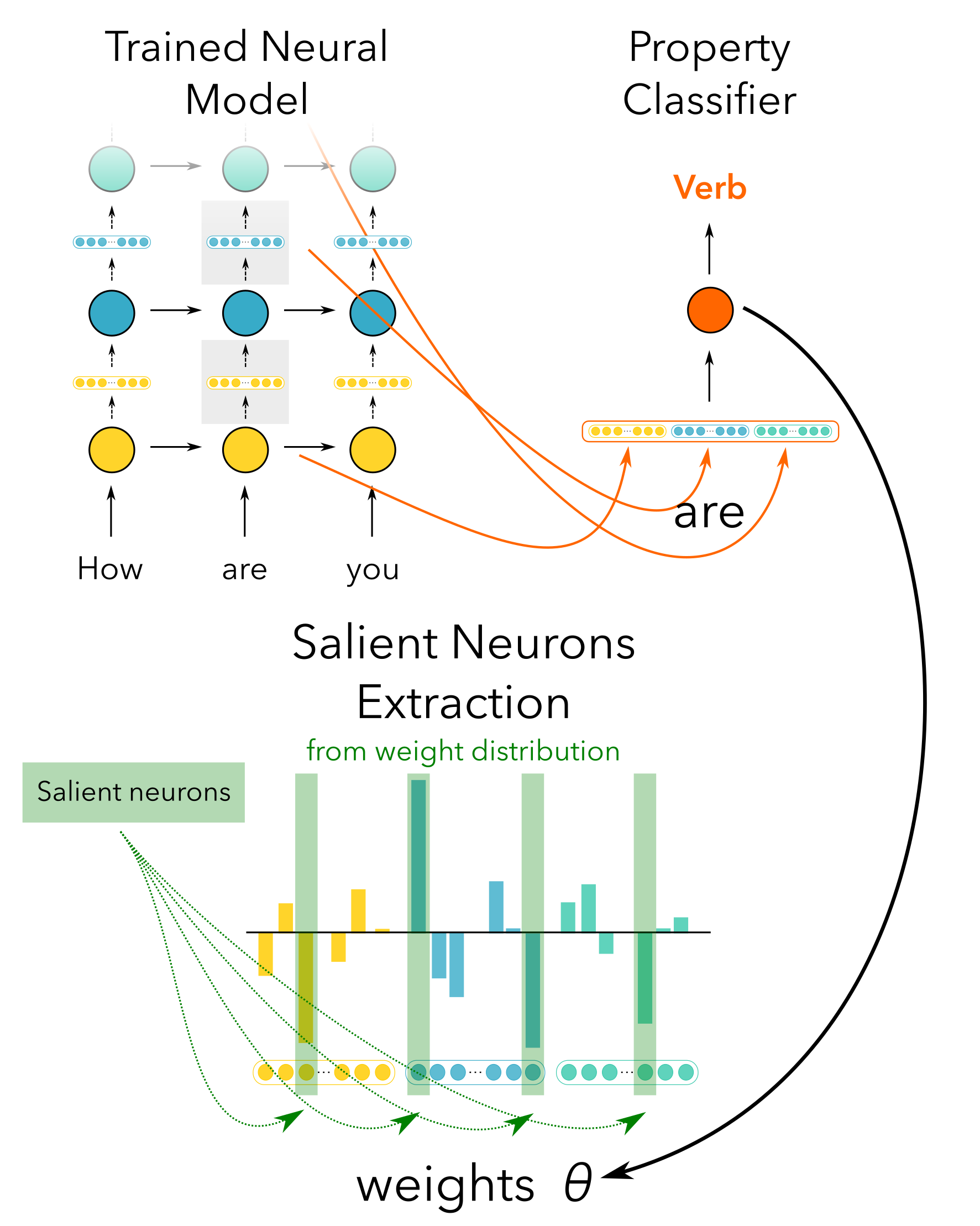}
	\caption{Linguistic Correlation Analysis: 
    Extract neuron activations from a trained model, train a classifier and use weights of the classifier to extract salient neurons.}
	\label{fig:supervised-method}
\end{figure}

Consider a classification task where the goal is to predict a property $\llll$ in a property set $\propP$\footnote{A property could be a part-of-speech tag such as {\tt verb}, or a semantic entity such as {\tt event}, or the position of a word in a sentence. A set of properties combined constitutes a task such as POS or semantic tagging. 
} that we believe is 
intrinsically learned in the model $\modelM$, for example word-structure (morphology) or semantic information in an NMT model. Our goal is to identify neurons in $\modelM$ that are salient for the property $\llll \in \propP$ being considered. We assume that we have supervision for the task in the form of labeled examples $\{\xx_i, \llll_i\}$ where $\xx_i$ is the $i$-th word, having a property $\llll_i \in \propP$. Given this labeled training data, we first extract neuron activations $\zz_i$ from the model $\modelM$ for every input word $\xx_i$. For instance, this may be done by running the NMT encoder on the sentence and recording neuron activations for each word.

We then train a logistic regression classifier on the $\{\zz_i,\llll_i\}$ pairs using the cross-entropy loss. We opt to train a linear model because of its explanability; the learned weights can be queried directly to get a measure of the importance of each neuron in $\zz_i$. From a performance point of view, earlier work has also shown that non-linear models present similar trends as of linear models in analyzing representations of neural models \cite{qian-qiu-huang:2016:P16-11,belinkov:2017:acl}. In order to increase interpretability and to encourage feature ranking in the classification process, we use elastic net regularization \cite{Zou05regularizationand} as an additional loss term. Formally, the model is trained by minimizing the following loss function:

\begin{equation}
\mathcal{L}(\theta) = -\sum_i \log P_{\theta}(\llll_i | \xx_i) + \lambda_1 \|\theta\|_1 + \lambda_2 \|\theta\|^2_2 \nonumber
\end{equation}

\noindent where $P_{\theta}(l | \xx_i) = \frac{\exp (\theta_l \cdot \zz_i)}{\sum_{l'} \exp (\theta_{l'} \cdot \zz_i)} $ is the probability that word $i$ is assigned label $l$. The weights \mbox{$\theta \in \reals^{D \times L}$} are learned with gradient descent. Here $D$ is the dimensionality of the latent representations $\zz_i$ and $L$ is the size of the label set for $\propP$. The overall process is illustrated in Figure \ref{fig:supervised-method}.

Elastic net regularization enjoys the sparsity effect as in Lasso regularization, which helps identify important individual neurons. At the same time, it takes groups of highly correlated features into account similar to Ridge regularization, avoiding the selection of only one feature 
as in Lasso regularization. This strikes a good balance between localization and distributivity. This is particularly useful in the case of analyzing neural networks where we hypothesize that the network consists of both individual focused neurons and a group of distributed neurons, depending on the property being learned. The regularization terms are controlled by hyper-parameters $\lambda_1$ and $\lambda_2$. We search for the best hyper-parameter values that maintain good accuracy while accomplishing the desired goal of selecting the salient neurons for a property, as described in the evaluation section.

\paragraph{Ranking Neurons:} Given the trained weights of the classifier \mbox{$\theta \in \reals^{D \times L}$}, we want to extract a ranking of the $D$ neurons in the model $\modelM$. For the label of interest $\llll \in \propP$, we sort the weights $\theta_{\llll} \in \reals^D$ by their absolute values in descending order. Hence the neuron with the highest corresponding absolute weight in $\theta_{\llll}$ 
appears at the top of our ranking. We consider the top $n$ neurons (for the individual property under consideration) that cumulatively contribute to some percentage of the total weight mass as \textit{salient neurons}. 
To extract a ranking of neurons w.r.t.\ all of the labels in $\propP$, we use an iterative process  described in Algorithm \ref{alg:ordering}. 
We start with a small percentage of the total weight mass and choose the most salient neurons for each label $\llll$, and increase this \% iteratively, adding newly discovered top neurons to our ordering. Hence, the salient neurons for each label $\llll$ will appear at the top of the ordering. The order in which the neurons are discovered indicates their importance to the property set $\propP$. 

\begin{algorithm}[t]
	\caption{Neuron Ranking Extraction Algorithm}
	\label{alg:ordering}
	\small
	\begin{algorithmic}[1]
		\State $ordering \gets$ []
		\Comment{{\footnotesize $ordering$ will store the neurons in order of decreasing importance}}
		\For{$p=1$ \textbf{to} $100$ \textbf{by} $\alpha$}
		\Comment{{\footnotesize $p$ is the percentage of the weight mass. We start with a very small value and incrementally move towards 100\%.}}
		\State $tnpt \gets$ \Call{GetTopNeuronsPerTag}{$\theta$, $p$}
		\Comment{{\footnotesize $tnpt$ contains the top neurons per tag using the threshold $p$}}
		\State $topNeurons \gets \bigcup\limits_{i=1}^{L} tnpt_	{i}$
		\State $newNeurons \gets topNeurons \setminus ordering $
		\State $ordering.append(newNeurons)$
		\EndFor
		\State \Return $ordering$
	\end{algorithmic}
\end{algorithm}

\subsection{Cross-model Correlation Analysis}
\label{sec:method-unsupervised}
The linguistic correlation analysis is useful for analyzing neurons given a certain property. Now, we present our Cross-model correlation method to identify neurons salient to the model $\modelM$ independent of any property. In essence, it ranks neurons according to their importance to the task the model $\modelM$ is trained on.
We hypothesize that salient neurons contain important information about the task and are shared across several models. To prove this, we train multiple models $\modelM_1,\dots,\modelM_N$ for the same task, using identical model settings but with differing training data and initialization. We then rank neurons in one of the models $\modelM_i$ by their best correlation coefficient with any neuron from a different model:
%
\begin{equation}
score(\modelM_{ij}) = \max_{\substack{1 \leq i' \leq N \\ 1 \leq j' \leq D \\ i \neq i'}} \rho(\modelM_{ij}, \modelM_{i'j'})  \nonumber
\end{equation}

\noindent where $\modelM_{ij}$ is the $j$-th neuron in the $i$-th model and $\rho(\modelM_{ij}, \modelM_{i'j'})$ is the Pearson correlation coefficient.\footnote{Here $\modelM_{ij} \in \reals^T$, corresponding to activations of neuron $j$ in model $i$, over an evaluation set of size $T$ words.}  We then consider the top neurons in this ranking as the most salient neurons for the overall model.

\subsection{Evaluation using Neuron Ablation}
Given the list of neurons from a trained model $\modelM$, we evaluate the rankings by challenging their presence in the network. We clamp the value of a subset of neurons to zero as in~\cite{s.2018on} and observe the degradation in performance, reflecting how much the network is dependent on them. 
Our hypothesis is that an ablation of the most important neurons should cause a larger drop in performance compared to the least important neurons. 
We apply ablation to both the classifier (to evaluate property-specific rankings) and the original model $\modelM$ (to evaluate model-level rankings).


\paragraph{Ablation in Classification} 
Given a trained classification model, we keep N\% top or bottom neurons and set the activation values of all other neurons to zero in the test set. We then reevaluate the performance of the already trained classifier. We expect to see low performance (prediction accuracy) when using only the bottom neurons versus using only the top neurons. We also retrain the classifier with only the selected N\% neurons. This serves multiple purposes: i) it confirms the results from the zeroing-out method, ii) it shows that much of the performance can be regained using the selected neurons, and iii) it facilitates the analysis of how distributed a particular property is across the network. 


\paragraph{Ablation in Neural Model $\modelM$:}
Here, we want to evaluate our rankings of neurons with respect to the model $\modelM$.
Given a ranked list of neurons, we incrementally zero-out N\% of the neurons starting from top or bottom and report the drop in performance in terms of BLEU scores (for NMT) or perplexity (for NLM). 



\begin{table}[t]
	\footnotesize
	\centering
	\begin{tabular}{l|cc|cc|cc}
		\toprule
		& \multicolumn{2}{c|}{French}  & \multicolumn{2}{c|}{English} & \multicolumn{2}{c}{German} \\
        \midrule
		& POS & Morph &  POS & SEM & POS & Morph \\
		\midrule
		MAJ & 92.8 & 89.5 & 91.6 & 84.2 & 89.3 & 83.7 \\
		\midrule
		NMT & 93.2 & 88.0 & 93.5 & 90.1 & 93.6 & 87.3  \\
		NLM & 92.4 & 90.1 & 92.9 & 86.0 & 92.3 & 86.5 \\
		\bottomrule
	\end{tabular}
	
	\caption{Classifier accuracy when trained on 
    activations of NMT and NLM models. MAJ: local majority baseline.}
	\label{tab:classifier_acc}
\end{table}

\section{Experimental Settings}
\paragraph{Neural Models:}
We experimented with two architectures: NMT based on sequence-to-sequence learning with attention \cite{bahdanau2014neural} and an LSTM based NLM~\cite{hochreiter1997long}.\footnote{We focus on standard architectures for these tasks and leave exploration of recent variants such as the Transformer~\cite{NIPS2017_7181} or QRNN~\cite{bradbury2016quasi} for future work.} We trained a 2-layer bidirectional NMT model with 500-dimensional word embeddings and LSTM states. 
The system is trained for 20 epochs, and the model with the best development loss is used for the experiments. We follow similar settings to train a unidirectional NLM model. 

\paragraph{Data:} 
We experimented with English$\leftrightarrow$French (EN$\leftrightarrow$FR) and German$\rightarrow$English (DE$\rightarrow$EN) language pairs. We 
used a subset of 2 million sentences from the United Nations multi-parallel corpus \cite{ZIEMSKI16.1195} for EN$\leftrightarrow$FR and from the data made available for the IWSLT campaign \cite{cettolo2014report} for DE$\rightarrow$EN. We split the 
parallel data for each language pair into three equal subsets 
to train three different models. For language models, we used the source side of the parallel corpora.



\paragraph{Language Properties:}
We evaluated our linguistic correlation method by selecting standard tasks of part-of-speech (POS), morphological and semantic tagging. 
The former two capture word structure in a language and the latter captures its nuanced meaning. Additionally we considered some general properties, such as the position of words in a sentence and predicting a \emph{months of year} tag.

%


\paragraph{Classifier Data:}
We used 20k 
source-side sentences, randomly extracted from the MT training data, for training the classifier, and 4k sentences in the official test sets for testing.
We tagged these sentences with standard taggers for the different properties; 
the details of these taggers 
can be found in the supplementary material.

\section{Evaluation}
In this section, we present the evaluation of 
our techniques: 

\subsection{Linguistic Correlation Analysis}
\paragraph{Classifier Performance:} We first evaluate the classifier performance 
to ensure that the learned weights are actually meaningful for further analysis and ranking extraction. The classifiers were trained using 
the activations of already trained neural models (NLM 
and NMT 
encoder\footnote{We limit ourselves to encoder activations for simplicity.}). 
Table \ref{tab:classifier_acc} shows accuracy of the classifiers trained for different language pairs and tasks on a blind test set.  The classifiers achieve higher accuracies compared to the local majority baseline\footnote{Selecting the most frequent tag for each word and the most frequent global tag for the unknown words.} (MAJ) in all cases, except for French (POS:NLM). The overall accuracy trend shows that the neurons possess sufficient information to predict these language properties.

\begin{figure}[t]
	\centering
	\includegraphics[width=\linewidth]{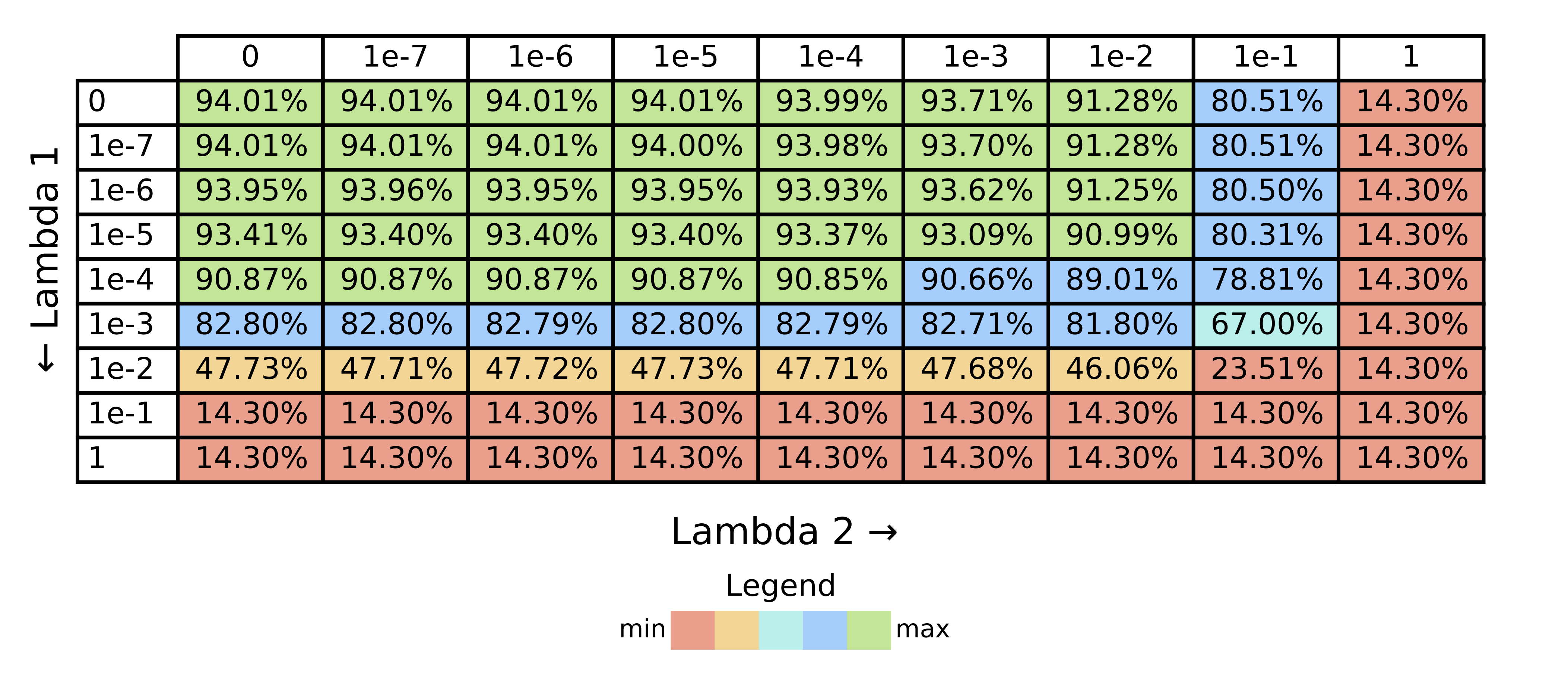} 
	\caption{Effect of various values of regularization on the overall accuracy of the English POS tagging task.}
	\label{fig:grid-search}
    
\end{figure}

Since we are using elastic net 
regularization, we need to tune the values for $\lambda_1$ and $\lambda_2$. The regularization controls the final ranking of the neurons directly: an increase in the value of $\lambda_1$ introduces further sparsity whereas higher values of $\lambda_2$ encourage selection of groups of correlated neurons. Our aim is to find a balance between selecting individual neurons and a group of neurons while maintaining the original accuracy of the classifier without any regularization ($\lambda_1$, $\lambda_2$ $=0$). Figure \ref{fig:grid-search} presents the results of a grid search over various regularization values on the English POS tagging task. The accuracy difference is minimal for $\lambda$ values under $1e^{-4}$. 
We selected a value of $1e^{-5}$ for both $\lambda_{1}$ and $\lambda_{2}$ and used the same for all the experiments. 

\begin{table}[t]
	\centering
	\resizebox{\linewidth}{!}{
	\begin{tabular}{p{0.2cm}|l|l||cc|cc|cc}
		\toprule
		\multicolumn{2}{l|}{} & & \multicolumn{6}{c}{Masking-out} \\
        \midrule
		\multicolumn{2}{l|}{Task}  & ALL & \multicolumn{2}{c|}{10\%} & \multicolumn{2}{c|}{15\%}  & \multicolumn{2}{c}{20\%} \\ 
		\multicolumn{2}{l|}{}  &  & Top & Bot & Top & Bot  &  Top & Bot \\
		\midrule
		\parbox[t]{0.1em}{\multirow{6}{*}{\rotatebox[origin=c]{90}{NMT}}} & FR (POS) &   93.2 & 63.2 & 23.8 & 73.0 & 24.8 & 79.4 & 24.9 \\
		&  EN (POS) &   93.5 & 69.8 & 15.8 & 78.3 & 17.9 & 84.1 & 21.5 \\
		&  EN (SEM) &   90.1 & 51.5 & 16.3 & 65.3 & 18.9 & 74.2 & 20.7 \\
		&  DE (POS) &   93.6 & 65.9 & 15.7 & 78.0 & 15.6 & 88.2 & 15.7 \\
		\midrule
		\parbox[t]{0.1em}{\multirow{6}{*}{\rotatebox[origin=c]{90}{NLM}}} &  FR (POS) &   92.4 & 41.6 & 23.8 & 53.6 & 23.8 & 59.6 & 24.0 \\
		& EN (POS) &   92.9 & 54.2 & 18.4 & 66.1 & 20.4 & 72.4 & 24.7 \\
		& EN (SEM) &   86.0 & 49.7 & 21.9 & 56.8 & 22.3 & 65.2 & 25.1 \\
		& DE (POS) &   92.3 & 39.7 & 16.7 & 51.7 & 16.7 & 67.2 & 16.9 \\
		\bottomrule
	\end{tabular}
    }
	\caption{
		Classification accuracy on different tasks using all neurons (ALL). 
		Masking-out: all except top/bottom N\% of neurons are masked when testing the trained classifier. 
	}
	\label{tab:classifier_ablation_mask_out}
\end{table}

\subsubsection{Neuron Ablation in the Classifier:}
\label{sec:classifier_ablation}
After training the classifier, we used Algorithm 1 to extract a ranked list of neurons with respect to each property set and ablated neurons in the classifier to verify rankings. We \emph{masked-out} all the activations (in the test set) except for the selected $N\%$ neurons and recomputed test accuracies. Table~\ref{tab:classifier_ablation_mask_out} summarizes the results.\footnote{
Similar trends were found in the morphological tagging results. Please see supplementary material if interested.} 
Compared to \texttt{ALL}, the classification accuracy drops drastically for both NMT and NLM. However, the performance is distinctly better in the case of keeping the top N\% neurons when compared to the bottom N\% neurons, showing that the ranking produced by the classifier is correct for the task at-hand. 

\begin{figure}[t]
	\centering
	\begin{subfigure}{1.0\linewidth}
		\includegraphics[width=\linewidth]{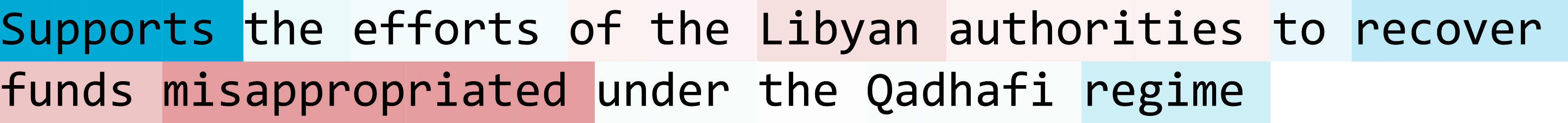}
		\caption{English Verb ({\tt\#1902})}
		\label{fig:en-verb}
	\end{subfigure}
    \begin{subfigure}{1.0\linewidth}
		\includegraphics[width=\linewidth]{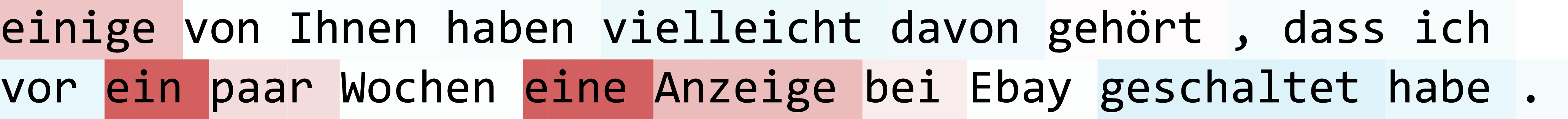}
		\caption{German Article ({\tt\#590})}
		\label{fig:de-article}
	\end{subfigure}
	\begin{subfigure}{1.0\linewidth}
		\includegraphics[width=\linewidth]{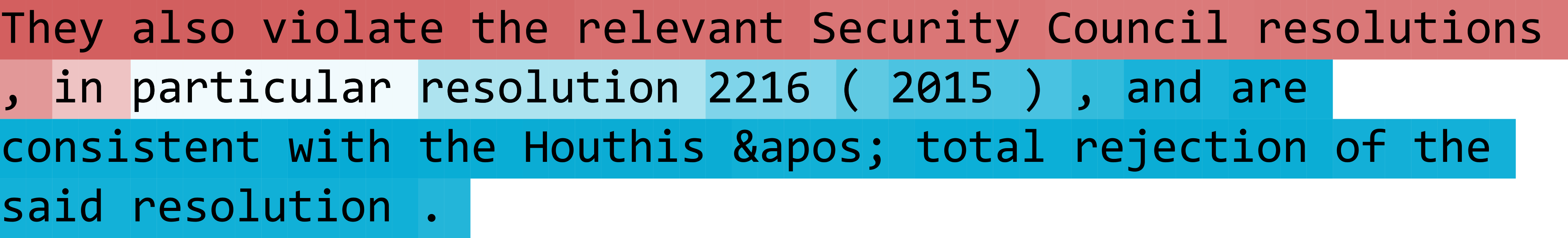}
		\caption{Position Neuron ({\tt\#1903})}
		\label{fig:en-position}
	\end{subfigure}
	\hfill
	\caption{Activations of 
    top neurons for specific properties 
    }
	\label{fig:vis}
\end{figure}


\paragraph{Visualizations:} have been used 
effectively to gain qualitative insights 
on analyzing neural networks  ~\cite{karpathy2015visualizing,kadar2016representation}. We used an in-house visualization tool \cite{neuroX:aaai19:demo} for qualitative evaluation of our rankings. Figure \ref{fig:vis} 
visualizes the activations of 
the top neurons for a few properties. 
It shows how single neurons can focus on very specific 
linguistic properties like \emph{verb} or \emph{article}. Neuron {\tt\#1902} focuses on two types of verbs (3$^{rd}$ person singular present-tense and past-tense) where it activates with a high positive value for the former (``Supports'') and high negative value for the latter (``misappropriated''). In the second example, the neuron is focused on German articles. Although our 
results are focused on linguistic tasks, the methodology is general for any property for which supervision can be created by labeling the data. For instance, we trained a classifier to predict \emph{position of the word}, i.e.,
identify if a given word is at the beginning, middle, or end of the sentence. As shown in Figure \ref{fig:vis}(a), the top neuron identified 
by this classifier activates with high negative value at the beginning (red), 
moves to zero in the middle (white), and gets a high positive value at the end of the sentence (blue). 
Another 
way to visualize is to look at the top words that activate a given neuron. Table \ref{tab:top_words} shows a few examples of neurons with their respective top 10 words. Neuron {\tt\#1925} is focused on the name of months. 
Neuron {\tt\#1960} is learning negation and Neuron {\tt\#1590} activates when a word is a number. 
These word lists give us quick insights into the property the neuron has learned to focus on, and allows us to interpret arbitrary neurons in a given network. 

\begin{table}[t]
	\footnotesize
	\centering
	\begin{tabularx}{\linewidth}{|p{1.7cm}|X|}
		\hline
		Neuron & Top 10 words \\
		\hline
		{\tt\#1925} \newline (Month) & August, July, January, September, October, presidential, April, May, February, December \\ 
        \hline
		{\tt\#1960} \newline (Negation) & no, No, not, nothing, nor, neither, or, none, whether, appeal \\ 
        \hline
        {\tt\#1590} \newline (Cardinality) & 50, 10, 51, 61, 47, 37, 48, 33, 43, 49 \\
		
        \hline
	\end{tabularx}
	\caption{Ranked list of words for some individual neurons in the EN-FR model.}
	\label{tab:top_words}
\end{table}

\subsection{Cross-model Correlation Analysis}
The Cross-model correlation analysis method ranks the list of neurons based on correlation among several models. In the following, we evaluate the rankings produced by the method by ablating the neurons in the original model $\modelM$. 

\begin{figure}[t]
	\centering
	\includegraphics[width=\linewidth]{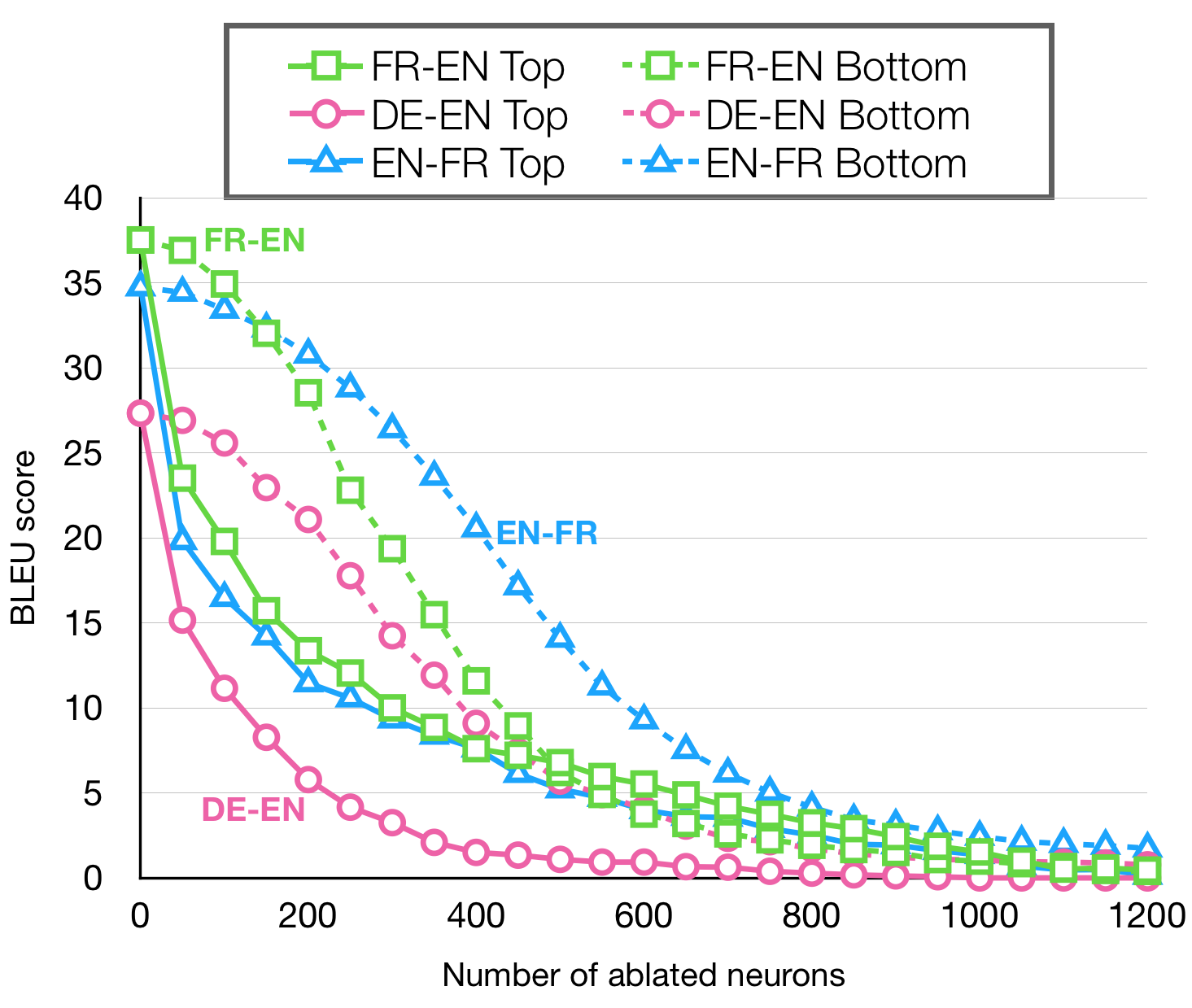} 
	\caption{Effect of neuron ablation on translation performance (BLEU) when removing the top or bottom neurons based on Cross-Correlation analysis ordering.}
	\label{fig:unsupervised-nmt-ablation}
\end{figure}

\paragraph{Neuron Ablation  in Model $\modelM$:} We incrementally ablate top/bottom neurons from the ranking and report the drop in performance of the NMT model. Figure \ref{fig:unsupervised-nmt-ablation} shows the effect of ablation on translation quality (BLEU).
For all languages, ablating neurons from top to bottom (solid curves) causes a significant early drop in performance  compared to ablating neurons in the reverse order (dotted curves). This validates the ranking identified by our method. Ablating just the top 50 neurons (2.5\%) 
leads to drops of 15-20 BLEU points, while  the bottom 50 neurons hurt the 
performance by 
only 0.5 BLEU points. 

\begin{figure}[ht]
	\centering
	\includegraphics[width=\linewidth]{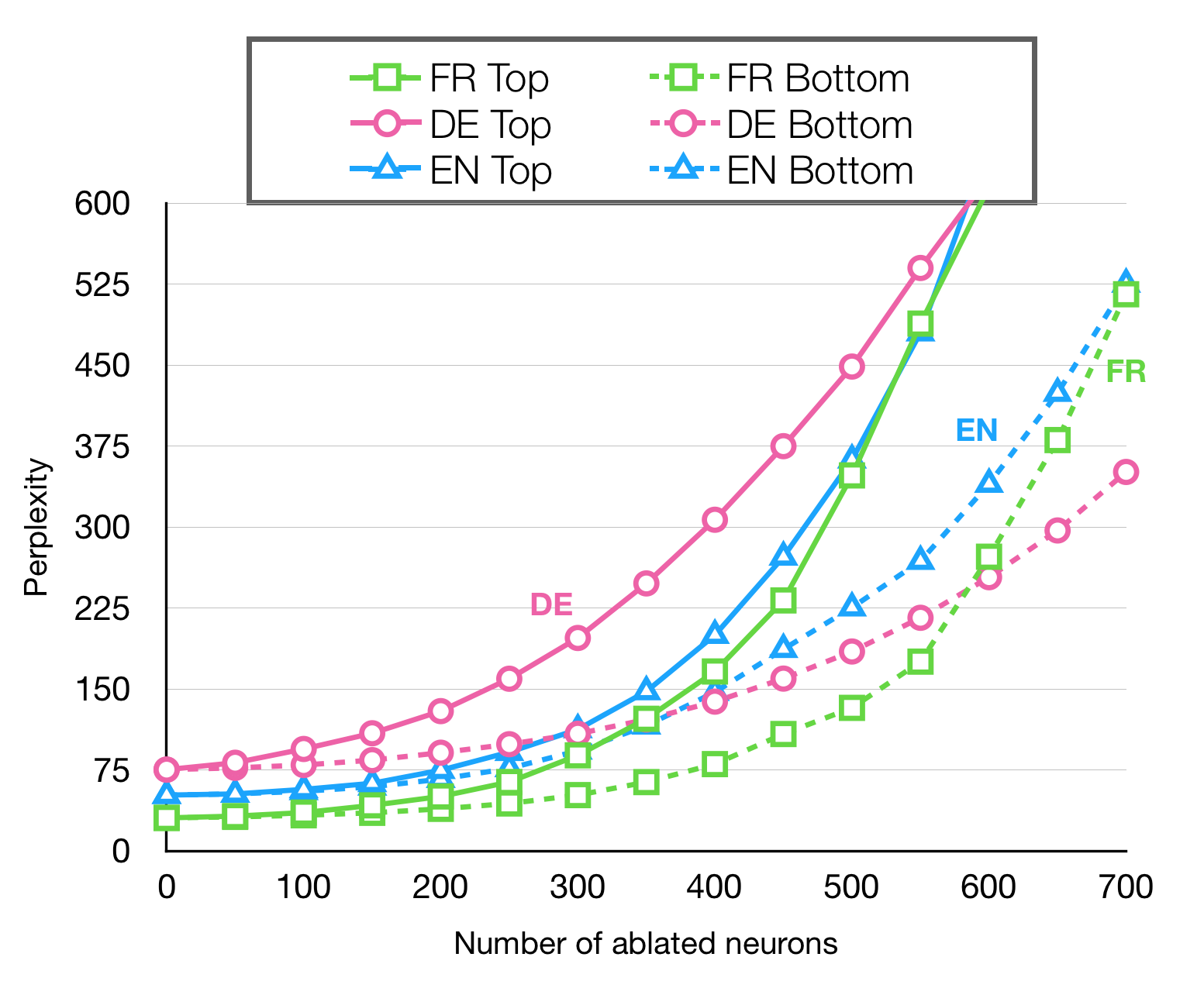} 
	\caption{Effect of neuron ablation on perplexity when erasing from the top and bottom of the Cross-correlation ordering from the NLM}
	\label{fig:unsupervised-nlm-ablation}
\end{figure}

\paragraph{Neuron ablation in NLM:}
Figure \ref{fig:unsupervised-nlm-ablation} presents the results of ablating neurons of NLM in the order defined by the Cross-model Correlation Analysis method. 
The trend found in the NMT results is also observed here, i.e. the increase in perplexity (degradation in language model quality) is significantly higher when erasing the top neurons (solid lines) as compared to when ablating the bottom neurons (dotted lines).

\begin{figure}[ht]
	\centering
	\includegraphics[width=0.9\linewidth]{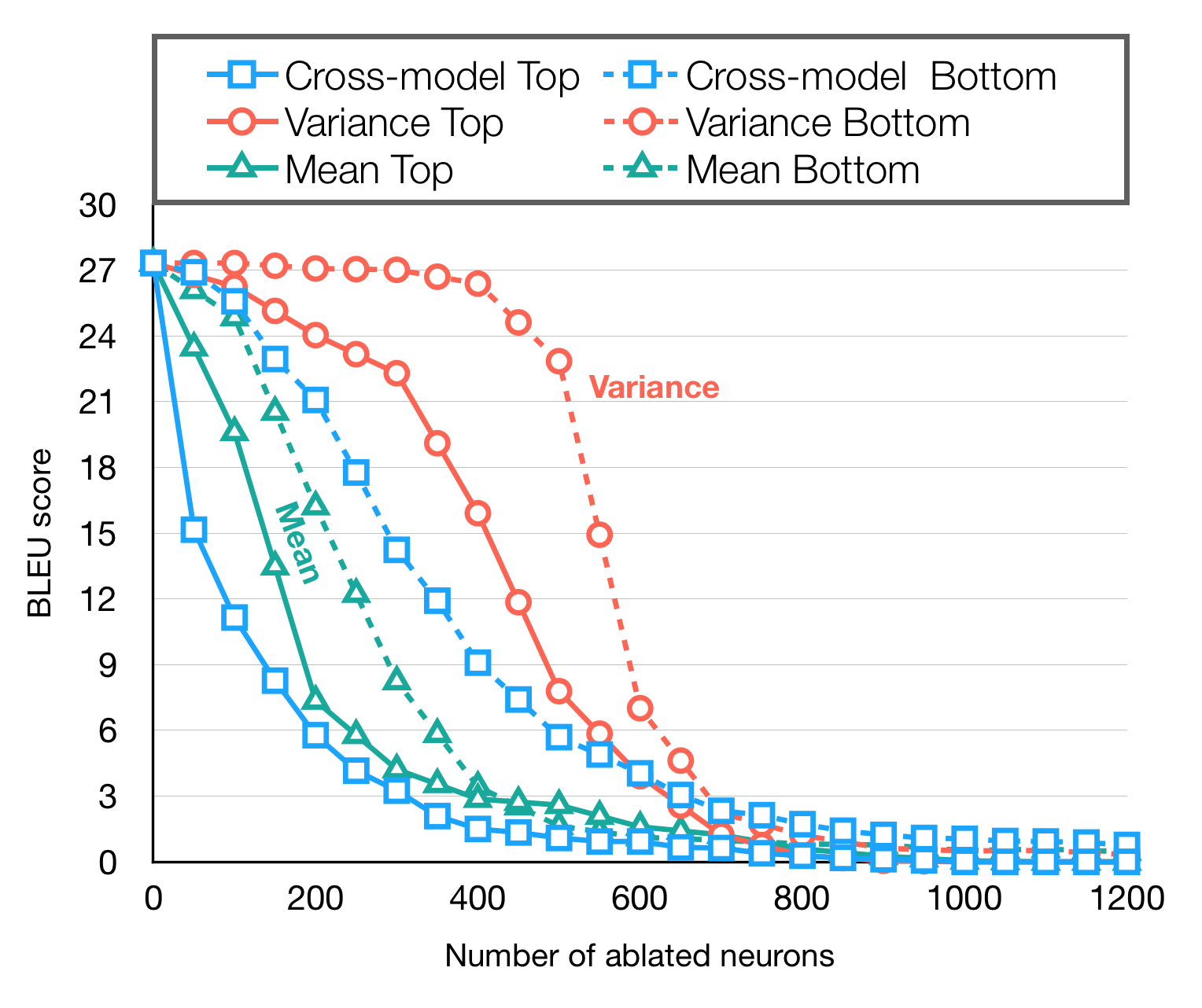}
	\caption{Cross-model ranking compared with single model statistics in DE-EN model. Variance is the ranking based on high variance to low variance. Mean is the ranking from high to low distance from mean.}
	\label{fig:single-model-analysis}
\end{figure}

\paragraph{Extracting Neurons based on a Single Model:} Recall that our Cross-model method requires multiple instances of the model to extract neuron rankings. In an effort to probe whether one instance of the model can sufficiently extract similar rankings, we tried several methods that ranked neurons of an individual model based on
i) variance, and ii) distance from mean (high to low), and compared these with the ranking produced by our method. We found less than 10\% overlap among the top 50 neurons of the Cross-model ranking and the single model rankings. On ablating the neurons based on several ranking methods, we found 
the NMT models 
to be most sensitive to the Cross-model ranking. Less damage was done when neurons were ablated using rankings based on variance and distance from mean in both directions, high-to-low and low-to-high (See Figure \ref{fig:single-model-analysis}). This supports our claim that the Cross-model ranking identifies the most salient neurons of the model.  

\begin{figure}[t]
	\centering
	\includegraphics[width=0.9\linewidth]{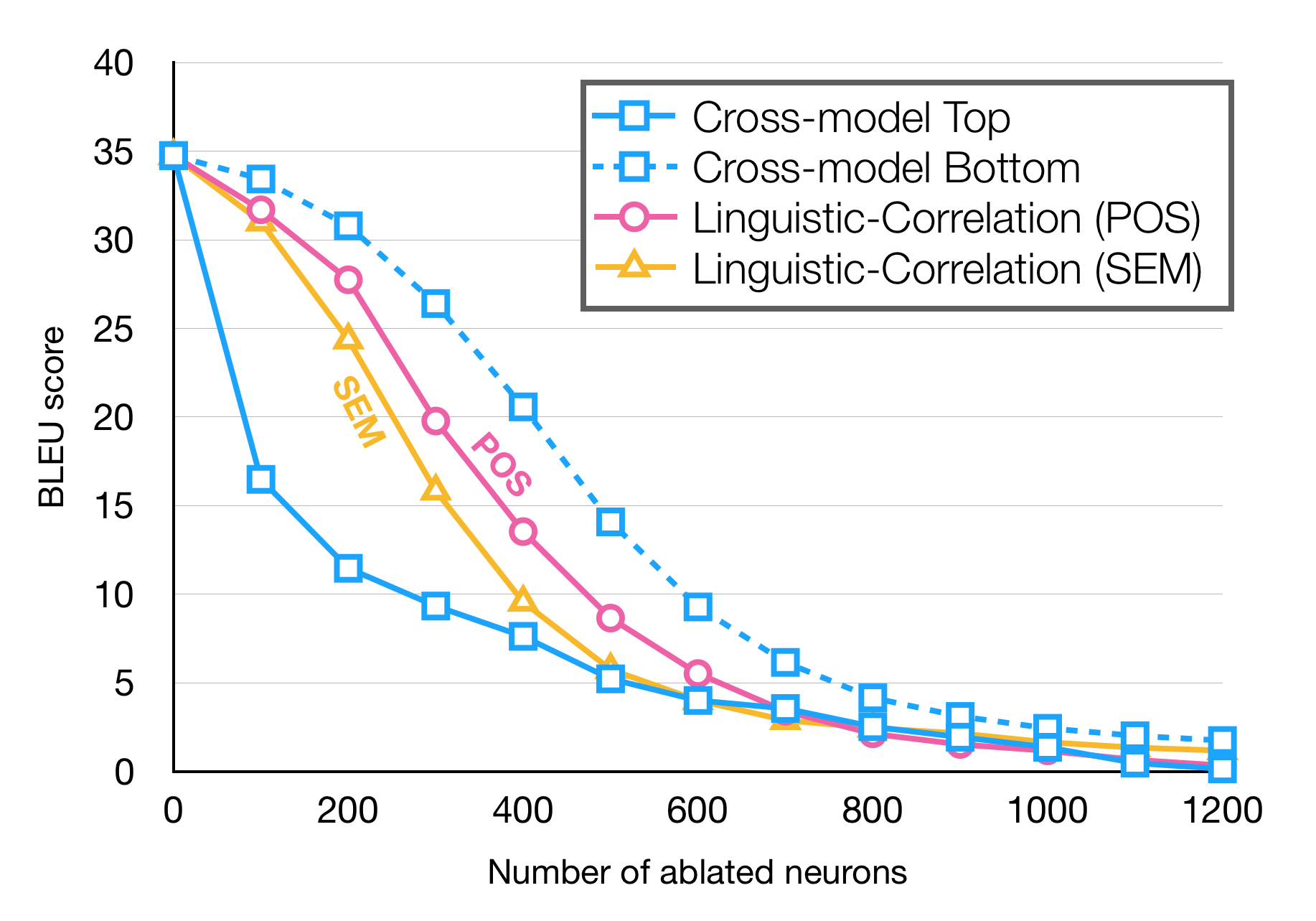} 
	\caption{
		Effect on translation when ablating neurons in the order determined by both methods on the EN-FR model
	}
	\label{fig:en-fr-unsupervised-supervised-ablation}
\end{figure}

\paragraph{Comparison with Linguistic Correlation Method:} Are the neurons discovered by the linguistic correlation method important for the actual model as well? Figure \ref{fig:en-fr-unsupervised-supervised-ablation} shows the effect on translation when ablating neurons in ranking order determined by English POS and semantic (SEM) tagging, 
as well as top/bottom Cross-model orderings. As expected, the linguistic correlation rankings are limited to the auxiliary task and may not result in the most salient neurons for the actual 
task (machine translation in this case); ablating according to task-specific ordering hurts less than ablating by (top-to-bottom) Cross-model ordering. However, in both cases, degradation in translation quality is 
worse than ablating by bottom-to-top Cross-model ordering. Comparing SEM with POS, it turns out that NMT is slightly more sensitive to neurons focused on semantics than POS.


\section{Analysis and Discussion}
\label{sec:neuron_analysis} 

The rankings produced by 
the linguistic correlation and cross-correlation analysis methods give a sense of the most important neurons for 
an auxiliary task or the overall model. 
We now dive into 
neuron analysis based on these rankings. 
\begin{figure}[t]
	\centering
	\includegraphics[width=\linewidth]{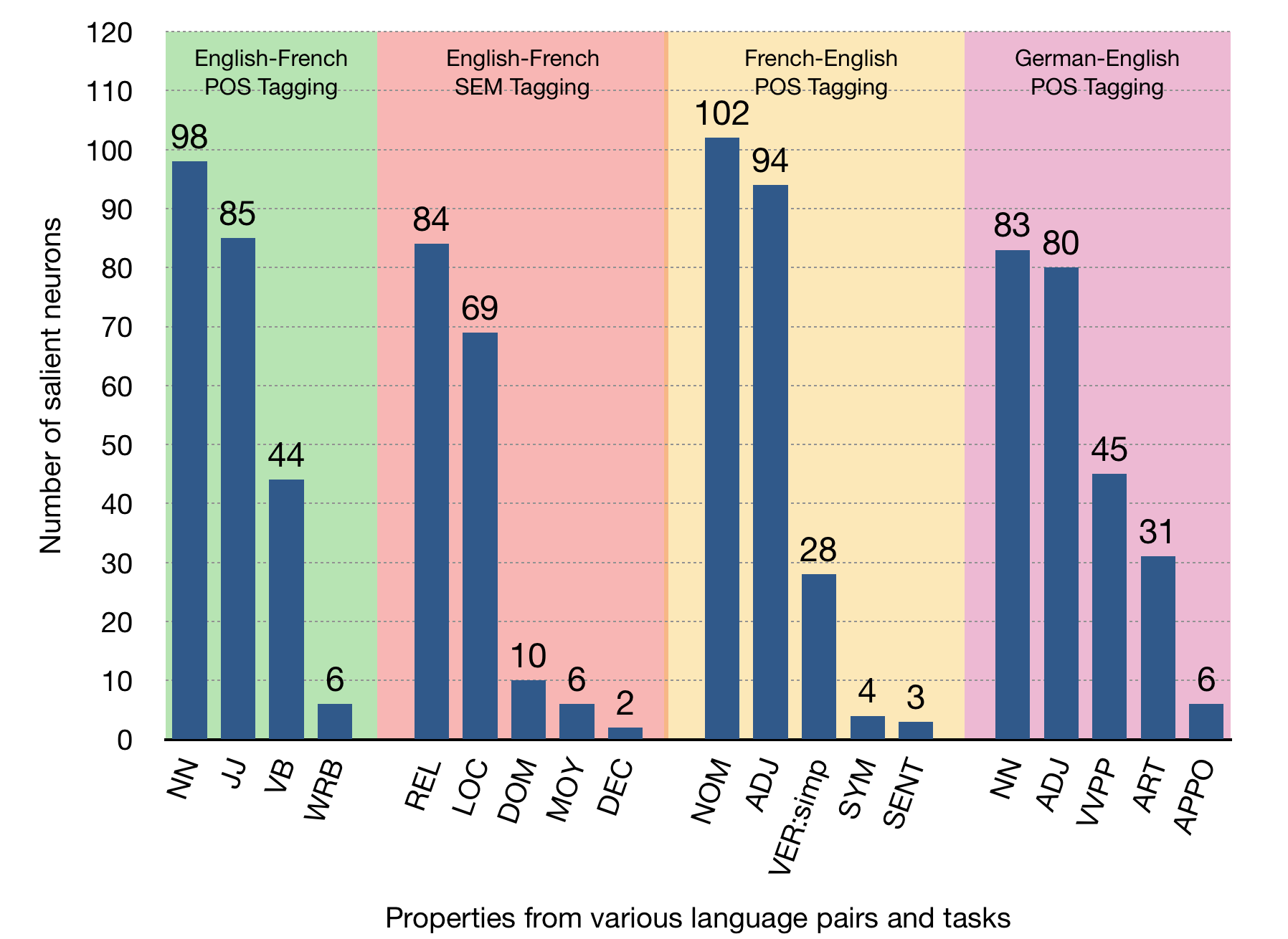}
	\caption{focused versus distributed tags: NN/NOM = Noun, JJ/ADJ = Adjective, VB = Verb, WRB = WH-Adverb, REL = relation, LOC = Location, DOM = Day of Month, MOY = Month of Year, DEC = Decade, VER:simp = Verb simple past, SENT = Full stop, VVPP = Participle Perfect, ART = Article, APPO = Post-position
	}
	\label{fig:task-distribution}
\end{figure}

\paragraph{Focused versus Distributed Neurons:}
Recall that our linguistic-correlation method provides an overall ranking w.r.t.\ a property set (POS/SEM tagging), 
and also for each individual property as described in the \textit{Methodology} section. Here, we look at the number of salient neurons (extracted from the NMT models) for several different linguistic properties,\footnote{We choose salient neurons for each label by selecting the top neurons that cumulatively represent 25\% of the total weight mass.} as shown in Figure \ref{fig:task-distribution}. For example, in open-class categories such as nouns (\texttt{NN/NOM}), verbs (\texttt{VB/VER.simp/VVPP}) and adjectives (\texttt{JJ/ADJ}), the information is distributed across several dozen neurons. In comparison, categories such as end of sentence marker (\texttt{SENT}) or WH-Adverbs (\texttt{WRB}) and post-positions (\texttt{APPO} in German) required fewer than 10 neurons. We observed similar trend in the semantic tags: information about closed-class categories such as months of year (\texttt{MOY}) is localized in just a couple of neurons. In contrast, an open category like location (\texttt{LOC}) is very distributed.

\paragraph{Shared Neurons within and across Properties:}
Since some information is distributed across the 
network, we expect to see some neurons that are common across various properties, 
and others that are unique to certain properties. To investigate this, we intersect top ranked neurons coming from two different properties.
Some of these comparisons are 
interesting. For instance, we found some common neurons across all forms of 
adjectives, but some neurons specifically designated to specialized adjectives (e.g., comparative (\texttt{JJR}) and superlative (\texttt{JJS}) adjectives).  Similarly across tasks (POS vs.\ Morph), 
we found multiple neurons targeting different verb forms (\texttt{V--F3s} and \texttt{V--F3p} ,  Verb Future 3$^{rd}$ person singular and plural) in the fine-grained morphological tagging that are aligned with a single neuron targeting the future tense verb 
tag (\texttt{VER:futu}) in  POS tagging. This demonstrates that model recognizes a hierarchy of linguistic properties and distributes neurons based on it.
\paragraph{Retraining Classifier with the Selected Neurons:}
In the evaluation section for our linguistic-correlation classifier, we masked-out a majority of the neurons and compared the accuracy trends to confirm our ranking. 
An alternative to analyze 
is to retrain the classifier with the top or bottom N\% neurons alone. Table \ref{tab:classifier_ablation_retrain} shows the results after retraining. 
There are several points to note here: i) training the classifier using top neurons performs consistently better than using bottom neurons, reinforcing our previous finding. 
ii) The classifier is able to regain performance substantially (compared to \texttt{ALL}),  even using only 10\% neurons. 
iii) 
Using the bottom N\% neurons also restores performance (although not as much as using the top neurons).  This shows that the information is distributed across neurons. However, the distribution is not uniform, which results in a large difference between training using top and bottom neurons (i.e., the information distribution is skewed towards the top neurons as expected). 
Notably, using only 20\% of the top neurons, the classifier is able to regain much of the performance drop in most of the cases. This 
finding entails that our method 
could be useful for model distillation purposes.

\paragraph{Cross-model Correlation Ranking:}
Analyzing the top neurons identified by our Cross-model correlation method, 
we found 
several 
neurons corresponding to the position of the word in a sentence. Word position has been previously found to be an important property in NMT~\cite{D16-1248}. The fact that our method ranks position neurons among the top ranking neurons 
shows its efficacy. 
We also observed that the top position neurons identified by our Linguistic Correlation method are the same as identified by the Cross-model correlation method.
Lastly, we 
found that some of the remaining top Cross-model neurons correspond to fundamental 
structural properties in a sentence, like relations, conjunctions, determiners and punctuations. 

\paragraph{Comparing NMT vs.\ NLM:}
There is substantially a large performance difference between top and bottom neurons 
(Refer to Table \ref{tab:classifier_ablation_retrain}). For example, 
averaged over all properties, the top 10\% NMT neurons are 12.8\% (absolute)  better accuracy than the bottom 10\% neurons, while the top 10\% NLM neurons are 25.5\% better than the bottom 10\% neurons. 
We speculate
that NMT model distributes the information 
more, compared to the NLM model. However, this could be an artifact of the difference in the architecture of NLM (unidirectional) and NMT (bidirectional).  

\begin{table}[t]
	\footnotesize
	\centering
	\resizebox{\linewidth}{!}{
	\begin{tabular}{p{0.2cm}|l|l||cc|cc|cc}
		\toprule
		\multicolumn{2}{l|}{} & & \multicolumn{6}{c}{Re-training}\\
        \midrule
		\multicolumn{2}{l|}{Task}  & ALL & \multicolumn{2}{c|}{10\%} & \multicolumn{2}{c|}{15\%}  & \multicolumn{2}{c}{20\%} \\ 
		\multicolumn{2}{l|}{}  &  & Top & Bot & Top & Bot  &  Top & Bot \\
		\midrule
		\parbox[t]{0.1em}{\multirow{6}{*}{\rotatebox[origin=c]{90}{NMT}}} & FR (POS) &   93.2 & 88.4 & 72.1 & 90.0 & 77.8 & 91.1 & 81.8 \\
		&  EN (POS) &   93.5 & 89.1 & 80.6 & 90.5 & 84.8 & 91.2 & 87.2 \\
		&  EN (SEM) &   90.1 & 85.6 & 73.4 & 87.0 & 77.8 & 87.8 & 80.8 \\
		&  DE (POS) &   93.6 & 91.4 & 77.1 & 92.3 & 81.9 & 92.8 & 85.3 \\
		\midrule
		\parbox[t]{0.1em}{\multirow{6}{*}{\rotatebox[origin=c]{90}{NLM}}} &  FR (POS) &   92.4 & 83.7 & 61.8 & 86.2 & 71.7 & 87.8 & 77.4 \\
		& EN (POS) &   92.9 & 85.8 & 62.4 & 88.2 & 72.5 & 89.4 & 79.2 \\
		& EN (SEM) &   86.0 & 78.9 & 67.8 & 81.4 & 74.1 & 82.7 & 77.6 \\
		& DE (POS) &   92.3 & 87.2 & 41.7 & 89.6 & 67.0 & 90.4 & 76.5 \\
		\bottomrule
	\end{tabular}
	}
	\caption{
		Classification accuracy on different tasks using all neurons (ALL). 
		Re-training: only top/bottom N\% of neurons are kept and the classifier is retrained
	}
	\label{tab:classifier_ablation_retrain}
\end{table}

\section{Conclusion and Future Work}


We proposed two methods to extract salient neurons 
from a neural model with respect to an extrinsic task or the model itself. 
We demonstrated 
the accuracy of our rankings by performing a series of ablation experiments. 
Our \emph{Cross-model Correlation} method can potentially facilitate research on 
model distillation and neural architecture search,
as it pinpoints what is especially important for the model. Our \emph{Linguistic Correlation} method 
is primarily focused on trying to understand specific 
dimensions that are responsible for learning particular properties. This can be helpful for understanding and manipulating systems' behavior. 
In some preliminary experiments, we were able to successfully manipulate verb tense neurons and control whether the system generates output in present or past tense. Some details are presented in \cite{indivdualneuron:arxiv19}.
The source code for extraction and analysis of salient neurons is incorporated in the NeuroX toolkit \cite{neuroX:aaai19:demo} and is available on git.\footnote{\url{https://github.com/fdalvi/NeuroX}} 



\section*{Acknowledgments}
We thank Preslav Nakov and the anonymous reviewers for their useful suggestions on an earlier draft of this paper. This work was funded by  Qatar Computing Research Institute, HBKU as part of the collaboration with the MIT Computer Science and Artificial Intelligence Laboratory (CSAIL). 

\section*{Supplementary Material}

\paragraph{Language Property Data:} We annotated the date using Tree-Tagger 
for French POS tags, LoPar 
for German POS and morphological tags, and MXPOST 
for English POS tags. 
For the semantic (SEM) tagging task, we experiment with the lexical semantic task 
introduced by \cite{bjerva-plank-bos:2016:COLING}.\footnote{The annotated data is limited to English language only.}
We split the available annotated data into 42k sentences for training and 12k sentences for testing.

\paragraph{Results on Morphological Tags:} Table \ref{tab:classifier_ablation_mask_out_morph} shows the results for the classifier performance when masking out neurons for morphological tags. Table \ref{tab:classifier_ablation_retrain_morph} shows the results when the classifier is retrained with N\% of the neurons.
\begin{table}[h]
	\centering
	\resizebox{\linewidth}{!}{
	\begin{tabular}{p{0.2cm}|l|l||cc|cc|cc}
		\toprule
		\multicolumn{2}{l|}{} & & \multicolumn{6}{c}{Masking-out} \\
        \midrule
		\multicolumn{2}{l|}{Task}  & ALL & \multicolumn{2}{c|}{10\%} & \multicolumn{2}{c|}{15\%}  & \multicolumn{2}{c}{20\%} \\ 
		\multicolumn{2}{l|}{}  &  & Top & Bot & Top & Bot  &  Top & Bot \\
		\midrule
		\parbox[t]{0.1em}{\multirow{2}{*}{\rotatebox[origin=c]{90}{NMT}}} & FR (Morph) & 88.0 & 25.2 & 17.3 & 39.0 & 20.3 & 56.3 & 24.3 \\
		&  DE (Morph)& 87.3 & 21.8 & 15.7 & 33.3 & 20.8 & 53.2 & 29.3 \\
		\midrule
		\parbox[t]{0.1em}{\multirow{2}{*}{\rotatebox[origin=c]{90}{NLM}}} &  FR (Morph) & 90.1 & 36.3 & 13.9 & 45.1 & 15.5 & 58.4 & 19.0 \\
		& DE (Morph) & 86.5 & 24.2 & 10.7 & 40.7 & 13.0 & 52.8 & 19.2 \\
		\bottomrule
	\end{tabular}
    }
	\caption{
		Classification accuracy on morphological tags for French and German using all neurons (ALL).
		Masking-out: all except top/bottom N\% of neurons are masked when testing the trained classifier. 
	}
	\label{tab:classifier_ablation_mask_out_morph}
\end{table}

\begin{table}[h]
	\footnotesize
	\centering
	\resizebox{\linewidth}{!}{
	\begin{tabular}{p{0.2cm}|l|l||cc|cc|cc}
		\toprule
		\multicolumn{2}{l|}{} & & \multicolumn{6}{c}{Retraining}\\
        \midrule
		\multicolumn{2}{l|}{Task}  & ALL & \multicolumn{2}{c|}{10\%} & \multicolumn{2}{c|}{15\%}  & \multicolumn{2}{c}{20\%} \\ 
		\multicolumn{2}{l|}{}  &  & Top & Bot & Top & Bot  &  Top & Bot \\
		\midrule
		\parbox[t]{0.1em}{\multirow{2}{*}{\rotatebox[origin=c]{90}{NMT}}} & FR (Morph) & 88.0 & 73.5 & 65.8 & 78.0 & 71.6 & 80.6 & 75.1 \\
		&  DE (Morph)& 87.3  & 79.3 & 75.4 & 82.1 & 78.9 & 83.5 & 80.5 \\
		\midrule
		\parbox[t]{0.1em}{\multirow{2}{*}{\rotatebox[origin=c]{90}{NLM}}} & FR (Morph) & 90.1 & 79.5 & 61.6 & 82.5 & 70.3 & 84.9 & 75.7 \\
		& DE (Morph) & 86.5 & 78.3 & 66.1 & 81.6 & 72.4 & 83.0 & 77.1 \\
		\bottomrule
	\end{tabular}
	}
	\caption{
		Classification accuracy on morphological tags for French and German using all neurons (ALL). 
		Re-training: only top/bottom N\% of neurons are kept and the classifier is retrained
	}
	\label{tab:classifier_ablation_retrain_morph}
\end{table}

\bibliography{naaclhlt2018,acl2017,thesis}
\bibliographystyle{aaai}

\end{document}